\documentclass[lettersize,journal]{IEEEtran}
\usepackage{amsmath,amsfonts}
\usepackage{algorithmic}
\usepackage{algorithm}
\usepackage{array}
\usepackage[caption=false,font=normalsize,labelfont=sf,textfont=sf]{subfig}
\usepackage{textcomp}
\usepackage{stfloats}
\usepackage{url}
\usepackage{verbatim}
\usepackage{graphicx}
\usepackage{cite}
\usepackage{color}
\usepackage{booktabs}
\usepackage{multirow}
\usepackage{hyperref}

\usepackage[capitalize]{cleveref}
\crefname{section}{Sec.}{Secs.}
\Crefname{section}{Section}{Sections}
\Crefname{table}{Table}{Tables}
\crefname{table}{Tab.}{Tabs.}

\usepackage{balance}

\begin{document}

\title{Towards Unified Token Learning for Vision-Language Tracking}

\author{Yaozong Zheng, Bineng Zhong$^{*}$, Qihua Liang, Guorong Li, Rongrong Ji, Xianxian Li
\thanks{Yaozong Zheng, Bineng Zhong, Qihua Liang, and Xianxian Li are with the Key Laboratory of Education Blockchain and Intelligent Technology, Ministry of Education, Guangxi Normal University, Guilin 541004, China, and the Guangxi Key Laboratory of Multi-Source Information Mining and Security, Guangxi Normal University, Guilin 541004, China.}
\thanks{Guorong Li is with the School of Computer Science and Technology, Key Laboratory of Big Data Mining and Knowledge Management, University of Chinese Academy of Sciences, Beijing 100049, China.}
\thanks{Rongrong Ji is currently a Professor with Media Analytics and Computing Lab, Department of Artificial Intelligence, School of Informatics, Xiamen University, 361005, China.}
\thanks{Bineng Zhong is the corresponding author.}}

\markboth{Journal of \LaTeX\ Class Files,~Vol.~14, No.~8, August~2021}%
{Shell \MakeLowercase{\textit{et al.}}: A Sample Article Using IEEEtran.cls for IEEE Journals}


\maketitle

\begin{abstract}

   In this paper, we present a simple, flexible and effective vision-language (VL) tracking pipeline, termed \textbf{MMTrack}, which casts VL tracking as a token generation task.
   Traditional paradigms address VL tracking task indirectly with sophisticated prior designs, making them over-specialize on the features of specific architectures or mechanisms.
   In contrast, our proposed framework serializes language description and bounding box into a sequence of discrete tokens.
   In this new design paradigm, all token queries are required to perceive the desired target and directly predict spatial coordinates of the target in an auto-regressive manner. 
   The design without other prior modules avoids multiple sub-tasks learning and hand-designed loss functions, significantly reducing the complexity of VL tracking modeling and allowing our tracker to use a simple cross-entropy loss as unified optimization objective for VL tracking task.
   Extensive experiments on TNL2K, LaSOT, LaSOT$_{\rm{ext}}$ and OTB99-Lang benchmarks show that our approach achieves promising results, compared to other state-of-the-arts.
\end{abstract}


\section{Introduction}
\label{sec:intro}

   \IEEEPARstart{V}{ision}-Language (VL) Tracking is an emerging multi-modal task, involving the cross-modal understanding on video frames and natural language description.
   Compared with vision-only tracking, the main difference of VL tracking is that it tracks the target instance by joint a natural language specification and initial frame in a video, which has great application potential in the real world, i.e., human-computer interaction and autonomous driving.

   \begin{figure}[t]
      \centering
       \includegraphics[width=1\linewidth]{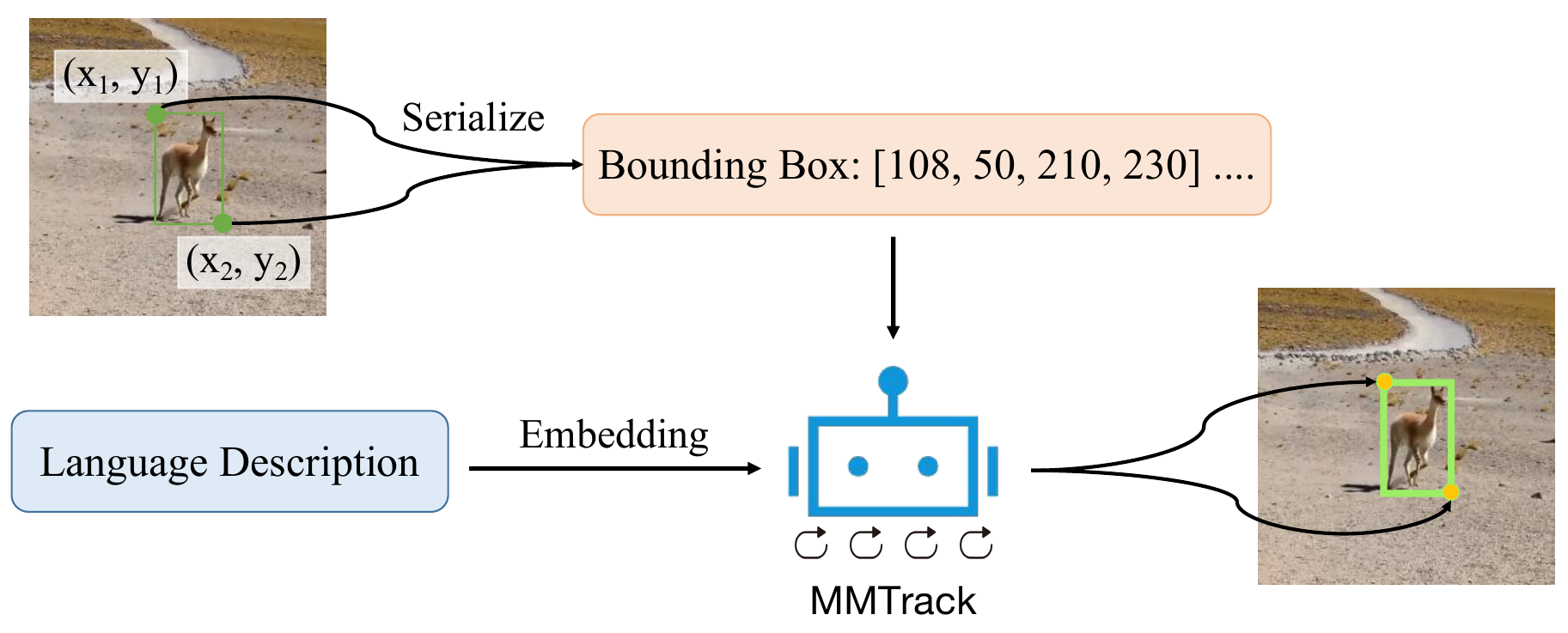}
    
       \caption{The process of MMTrack generating token sequences for target instance.
       Firstly, the spatial coordinates of the target are serialized into a sequence of discrete coordinate tokens and language token is embedded from natural language description.
       Then, MMTrack receives them as query inputs, and iteratively generates target sequences using an auto-regressive manner.
       }
       \label{fig:motivation}
    \end{figure}

   In the past few years, traditional VL tracking developments \cite{trackNL,DATrack,GTI,lstmtrack,NL-RPN,SNLT,tnl2k,VLT,cmtrack} solve this task indirectly by converting to grounding, retrieval or matching problem.
   Specifically, according to the different strategies of cross-modal fusion, three types of pipelines are commonly adopted:
   (1) \emph{Grounding-based VL tracking methods} \cite{GTI,tnl2k}. These methods introduce complicated hand-crafted components, i.e., AdaSwitcher \cite{tnl2k} and grounding network \cite{GTI}, that refer to a target instance in video frames from natural language description, then generate bounding box for the target instance.
   (2) \emph{Retrieval-based VL tracking methods} \cite{cmtrack}. These methods are commonly designed as a two-stage pipeline, with the first stage capturing candidate objects and the second stage using retrieval modules to construct cross-modal associations of proposals-text.
   (3) \emph{Matching-based VL tracking methods} \cite{NL-RPN,SNLT}. These methods incorporate information from vision and language modalities through the matching operators including natural-language region proposal network \cite{NL-RPN} and multi-level dynamic aggregation module \cite{SNLT}, and then utilize various conventional prediction heads to predict object bounding box.

   Despite good tracking results being produced, these efforts typically require expensive prior knowledge in term of network design.
   For example, some other VL trackers \cite{tnl2k,NL-RPN,cmtrack} are tailored to bounding boxes, and they use region proposal network (anchor-based mechanism) and ROI pooling to generate proposal instances for cross-modal fusion and alignment.
   In contrast, we discard these prior designs and let the neural network learn cross-modal representations about the target instance automatically, then the model simply relies on the learned unified VL features to generate useful target representations.

   On the other hand, previous algorithms \cite{tnl2k,GTI,cmtrack} also show that it is difficult to train a well-performing VL tracking model in a multi-task learning environment.
   One of the important reasons is that it is complicated to choose a training objective that is beneficial for vision-language understanding. 
   For instance, they often use various loss functions (such as triplet loss \cite{tripletloss}, focal loss \cite{focalloss}, GIoU loss \cite{giou}, and diversity loss \cite{diversityloss}) to supervise the proposed modules or tasks with specific functions. 
   Further, if a model wants to efficiently learn features for all types of tasks, the weights of multiple loss functions also need to be carefully tuned to fit sub-tasks, which makes them extremely difficult and has limited generalization capability.
   Therefore, these above studies are prone to produce sub-optimal solution.

   To simplify VL tracking modeling, we present a conceptually simple yet effective VL \textbf{M}ulti-\textbf{M}odal tracking pipeline, termed \textbf{MMTrack}.
   Specifically, we reformulate vision-language tracking as a token generation task that predicts bounding box in an auto-regressive manner. The core idea is that the language description and bounding box is serialized into a sequence of discrete coordinate tokens.
   Notably, we focus on a more challenging problem aimed at developing a unified VL multi-modal tracking framework considering both visual and linguistic representations, which is limited for unimodal trackers.
   We unify language and bounding box into multi-cues conditional queries, which are used as the coordinate prompt inputs of the multi-modal decoder, as shown in \cref{fig:motivation}. In this manner, all token queries can be uniquely referred to the target object in the video frame, and the high quality target instance information can be decoded by language and visual cues.
   In addition, due to the design without other prior modules, we avoid redundant sub-tasks learning and hand-designed loss functions, and only use a simple cross-entropy as a unified training objective. 
   
   The main contributions of this work are as follows.
    \begin{itemize}
    \item We reformulate vision-language tracking as a token generation task, and propose a novel pipeline to release the potential of VL multi-modal learning from a unified modeling perspective.
    
    \item The whole approach is simple and flexible, unifying language and bounding box as multi-cues token inputs. It avoids redundant sub-task learning and optimization objectives, and uses only cross-entropy as the unified training objective.
    
    \item Our approach achieves state-of-the-art tracking results on four VL benchmarks, suggesting that the proposed method can be a new baseline for VL tracking.
    \end{itemize}

\section{Related Work}
\label{sec:related_work}

\subsection{Visual Tracking}
   Recently, a large number of tracking algorithms have been proposed, which have greatly facilitated the development of visual tracking task.
   For instance, to build robust object appearance models, these studies \cite{SiamFC,SiamRPN,SiamRPN++,Siamban,Ocean,uncerTrack,transt,stark,SiamPIN,ostrack,seqtrack} explore more comprehensive and deeper interactions between template and search frame to better propagate visual features.
   
   They are able to perform well on visual benchmarks, but have a limited ability to extend the vision-language tracking task.
   This is primarily due to the fact that the tracking algorithms described above are specifically designed for vision-only tracking task, with lack flexibility in dealing with multi-modal data inputs such as video frames and language descriptions.
   In contrast to these approaches, our goal is to explore a unified multi-modal tracking pipeline with visual and language modalities.
   We employ transformer as the generic computation engine and formulate video frames and text inputs uniformly in the form of token sequences, so as to effectively capture their cross-modal relationships.

\subsection{Vision-Language Tracking}
   Vision-Language Tracking task requires the model to uniquely locate the state of the target object by the language description and initial frame in a video.
   \cite{trackNL} first proposes VL tracking task and designs a lingual specification attention network to extract description features, which facilitates the development of several subsequent studies.
   The current VL tracking approaches mainly follow three types of pipelines:
   (1) \emph{Grounding-based VL tracking methods}.
   \cite{GTI} decomposes VL tracking into three sub-tasks: Grounding, Tracking, and Integration.
   \cite{tnl2k} introduces a new TNL2K dataset and designs a visual grounding network based on a local-global search switcher strategy.
   (2) \emph{Retrieval-based VL tracking methods}.
   \cite{cmtrack} provides multiple sub-modules and loss functions to solve cross-modal semantic alignment problem.
   (3) \emph{Matching-based VL tracking methods}. 
   \cite{NL-RPN} and \cite{SNLT} introduce a natural-language region proposal network and a multi-level dynamic aggregation module to cross-modal fusion for VL tracking. 

   These above studies indirectly solve VL tracking task by designing the paradigms of independent modules and multiple sub-task learning.
   In contrast to these works, we avoid such complexities by treating VL tracking task as a token generation task, and propose a new baseline approach, called MMTrack, that aims to unlock the potential of vision-language tracking by learning unified multi-modal representation.

\subsection{Unified Modeling of Language and Vision}
   In recent years, unified learning of language and vision has become a popular research topic \cite{clip,OFA,UniT,MDETR,UniTAB}. 
   Specifically, CLIP \cite{clip} learns a robust vision-language model for zero-shot image classification by using a contrastive learning strategy, which can be easily transferred to various downstream tasks.
   UniT \cite{UniT} puts forward a unified transformer model, and designs multiple task-specific prediction heads to jointly learn multiple cross-domain tasks, and achieves surprising results.
   Based on the recent DETR \cite{DETR}, MDETR \cite{MDETR} proposes an end-to-end multi-modal detection baseline that can be combined with natural language understanding for object detection to achieve multi-modal reasoning.
   OFA \cite{OFA} designs a task-agnostic and modality-agnostic framework to unify a variety of cross-modal and unimodal tasks, including image generation, visual grounding, image captioning, and so on.
   Pix2seq \cite{pix2seq} solves object detection task using a language modeling way and achieves competitive results.

\begin{figure*}
\begin{center}
\includegraphics[width=1\linewidth]{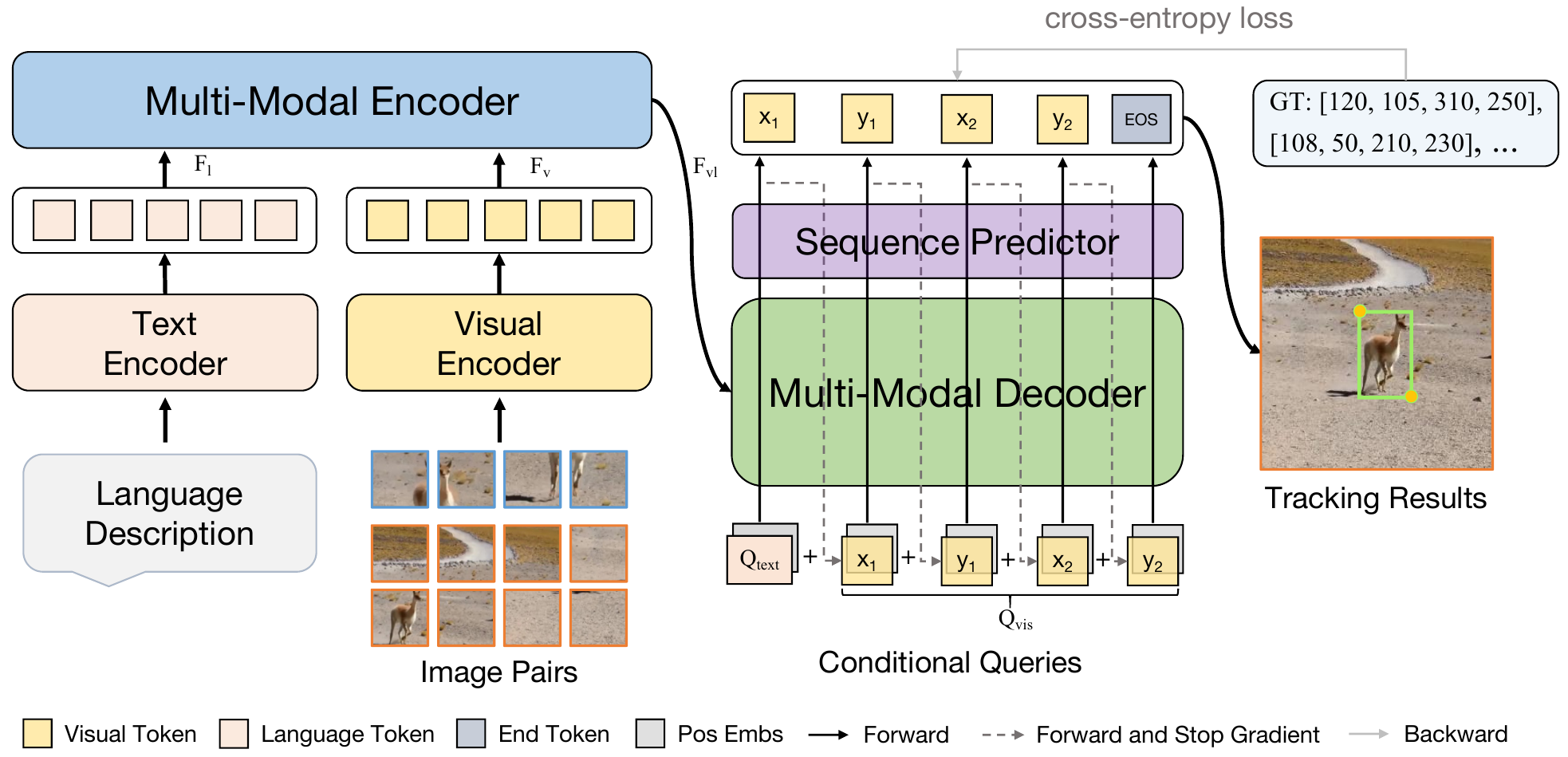}
\end{center}
\vspace{-3mm}
\caption{\textbf{MMTrack Framework Architecture.} (a) We extract linguistic and visual features using a text encoder and a visual encoder, respectively. (b) We feed two types of features into the multi-modal encoder for unified VL representation learning. (c) We formulate a unified set of conditional queries with guidance from the linguistic and bounding box cues. Next, we use all of the conditional queries and VL representation as inputs to the multi-modal decoder. The conditional queries learn to generate the target sequence with bounding box information from the VL representation in an auto-regressive fashion.}
\label{fig:framework}
\end{figure*}

\subsection{Discussion on the difference with sequence-based work}
   Inspired by the philosophy of pix2seq\cite{pix2seq}, we contribute a novel VL multi-modal tracking model for the tracking community.
   However, our work differs significantly from pix2seq in the following aspects: multi-modal modeling and sequence construction strategy.
   (1) Pix2seq is designed for object detection, while we creatively extend this concept to the multi-modal modeling of VL tracking task and design a multi-modal encoder for text-video understanding. This is limited for the unimodal pix2seq.
   (2) Pix2seq constructs long sequences from bounding boxes, categories, and noisy data. In contrast, we simplify the sequence construction strategy by relying solely on the bounding box and text information, avoiding the use of categories and other noise strategies. As a result, by preventing long and complex sequences, our method reduces the training load and is better adapted to the VL tracking task.

   On the other hand, SeqTrack \cite{seqtrack} is similar to our study, with the following notable differences:
   (1) SeqTrack is specifically designed for vision-only tracking task involving visual modality,  with lack flexibility in dealing with multi-modal data inputs. In contrast, vision-language tracking incorporates high-level semantic information to address the ambiguity of bounding boxes, resulting in more flexible, robust, and accurate tracking in practical applications. Specifically, our approach operates on multi-modal data as input and requires careful consideration of multi-modal fusion operations. the input of MMTrack is a triplet consisting of one language description and one template-search pair. Further, we utilize a text encoder, a visual encoder and a multi-modal encoder to achieve cross-modal understanding for VL tracking task.
   (2) The token sequence construction strategies are different. Despite both SeqTrack and our work using the construction strategy of short token sequences, we reduce the number of quantization bins and also consider incorporating language tokens derived from linguistic descriptions. This addition enhances the robustness of target localization in complex scenarios.

\section{Approach}
\label{sec:method}
   
   In this section, we introduce MMTrack, a unified multi-modal tracking framework for vision-language tracking, as shown in \cref{fig:framework}. 
   The MMTrack contains two inputs: image pairs and language description.
   To begin, we extract two types of features from language descriptions and video frames using a text encoder and a visual encoder, respectively.
   To speed up the computational efficiency, we use two linear layers to reduce the channel dimension of two types of features from $C$ to $d$.
   We then feed linguistic and visual features into the multi-modal encoder for unified VL representation learning, as developing a unified representation via fusion operations is the key to multi-modal learning.

   To construct conditional queries, we tokenize the text embedding and bounding box to produce multiple 1-dim token sequences, which are concatenated to obtain our conditional queries.
   Following that, the conditional queries and VL representation are fed into the multi-modal decoder. In an auto-regressive fashion, the conditional queries learn to generate the target sequence with bounding box information from the VL representation.
   
   Finally, we design a simple task-agnostic sequence head that can predict final tracking results directly.
   The overall pipeline of MMTrack is simple and flexible. We believe it is a meaningful step toward making vision-language tracking more generic and accessible.

\subsection{Encoders Across Modalities}
   \textbf{Visual Encoder.} 
   Modern visual encoders for extracting image features can be divided into two groups. One is convolutional neural networks \cite{resnet,googlenet} and the other is vision transformer architectures \cite{vit,CvT}.
   These visual encoders require perceiving and understanding the content of the input images.
   In our framework, we encode the input image pairs with a vision transformer, into a series of visual token sequences.
   Specifically, we apply a vanilla ViT \cite{vit} as the visual encoder, which receives an image pair as input containing a template frame $z \in \mathbb{R}^{3 \times H_z \times W_z}$ and a search frame $x \in \mathbb{R}^{3 \times H_x \times W_x}$.
   To start, the template and search frames are projected to the feature space and flattened to produce token embeddings using a patch embedding layer. Next, the token embeddings are concatenated and loaded into the transformer layers for feature extraction and relational modeling.

   \textbf{Linguistic Encoder.}
   In contrast to vision-only tracking, VL tracking requires tracking the target object in a video sequence using a language specification.
   We use RoBERTa \cite{RoBERTa}, a transformer encoder model that has been pretrained on large corpora.
   Similar to the visual encoder, we tokenize the input sentence into a sequence of text tokens. The token sequences then are fed into the RoBERTa language model to extract text embedding vectors.
   It is worth noting that linguistic feature extraction plays a crucial role in our pipeline, which is not only used for multi-modal fusion, but also for generating language query.

   \textbf{Multi-modal Encoder.}
   Previous VL trackers \cite{tnl2k,SNLT,cmtrack} typically used separate modules with sophisticated design to fuse visual and linguistic features, which obviously increases the complexity of multi-modal tracking modeling.
   Since transformer is able to propagate information sufficiently between text and visual features, which helps to obtain accurate VL features, we choose transformer as the general computing engine of VL tracking framework.
   After encoding two types of inputs including visual and language modalities, we use a shared multi-modal encoder for each source to perform fine-grained cross-modal understanding, so as to achieve unified VL representation learning.

   To be specific,
   given two types of encoded linguistic $f_{l} \in \mathbb{R}^{C \times N_{l}}$ and visual $f_{v} \in \mathbb{R}^{C \times N_{v}}$ features, we perform a pre-processing step in which two bottleneck layers are utilized to reduce the channel numbers of both features from $C$ to $d$, respectively.
   We then concatenate them and supply them into the shared multi-modal encoder to construct a reliable cross-modal interaction.
   The formulation is as follows:
   \begin{equation}
     F_{vl} = \delta \left( \phi(f_{v}, f_{l}) \right) \times \phi(f_{v}, f_{l})
     \label{eq:fusion}
   \end{equation}
   where $\phi$ denotes concatenate operation, $\delta$ represents multi-modal fusion function, and $F_{vl}$ is the learned unified VL representation.
   Notably, we use residual multiplication, which is beneficial for capturing rich semantic information in large-scale text-video data.
   We can observe that this unified modeling way not only simplifies the previous VL tracking procedure, but also achieves a deeper cross-modal understanding between text and video frames.

\subsection{Multi-modal Decoder}
   In the NLP community, language models \cite{GPT-3} typically use auto-regressive prediction to unify the output of various language tasks into a sequence-to-sequence format.
   Inspired by this philosophy, we design a multi-modal decoder with auto-regressive transformer to decode the target sequence for VL tracking.
   
\begin{figure}[t]
      \centering
       \includegraphics[width=1\linewidth]{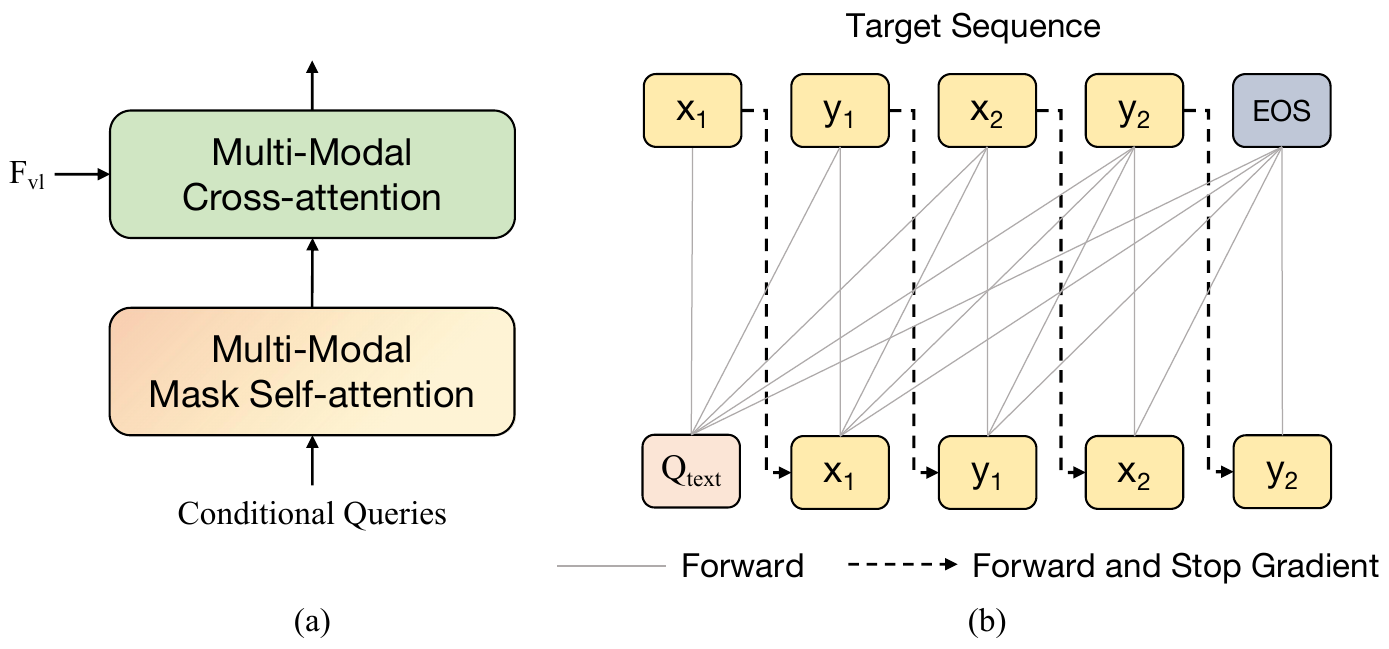}
       \caption{(a) The architecture of multi-modal decoder. (b) The multi-modal decoder receives linguistic and visual token sequences as query inputs, and iteratively generates target sequences using an auto-regressive manner.}
       \label{fig:decoder}
   \end{figure}

   Specifically, as shown in \cref{fig:decoder}(a), the proposed decoder consists of multiple multi-modal causal mask self-attention layers and multi-modal cross-attention layers.
   we first add \emph{sine} and \emph{learnable} position encoding ($P^{sine}$ and $P^{learn}$) to VL representation and input queries, respectively.
   Then, as shown in \cref{fig:decoder}(b), the self-attention layer equipped with a causal mask receives queries from previous input queries (token sequences), which guarantees that the output of each target sequence is solely dependent on its previous token sequences, thus upholding the fundamental principle of causality.
   Subsequently, the VL representation and input queries are fed into the multi-modal cross-attention layer to learn target sequence representation in a cross-interpretion manner.
   \begin{equation}
     T_{seq} = \sum_{j=1}^{L} D ( F_{vl} + P^{sine}, Q_{0:j-1} + P_{0:j-1}^{learn} )
     \label{eq:decoder}
   \end{equation}
   where 
   $D$ denotes multi-modal decode function, $Q_{0:j-1}$ represents the $0$-th to $j-1$-th consecutive input queries, $L$ is the target sequence length, and $T_{seq}$ denotes the decoded target sequence representation.

   Under this paradigm, different modalities can be propagated in the MMTrack with an encoder-decoder architecture.
   The multi-modal encoder is responsible for updating the VL features, whereas the multi-modal decoder predicts the discrete coordinate tokens in an auto-regressive fashion.
   As you can see, the whole pipeline is simple and flexible.

\subsection{Conditional Queries}
\label{sec:query}

   In VL tracking datasets \cite{tnl2k,lasot,lasot-ext,trackNL}, the target object in each video frame is represented as a bounding box and a language description.
   The bounding box describes the target object's appearance information, while the language description expresses the semantic information of the target object.
   To take full advantage of the different information from both cues, we propose conditional queries, i.e. a new query formulation based on vision and language modalities.
   Specifically, the conditional queries are formed by two parts: a language query $Q_{text}$ with text embedding and a vision query $Q_{vis}$ with box coordinates. 
   These queries are required to perceive the desired target from both modalities and directly predict spatial coordinates of the target in an auto-regressive manner.
   Benefiting from this, MMTrack avoids the design of additional independent modules, greatly reducing the complexity of multi-modal tracking modeling.

   \textbf{Query Construction with Multi-Modal Cues.}
   In comparison to prior works \cite{stark,ostrack,VLT,cmtrack}, the key design of MMTrack stems from the construction of query sequences.

   To construct language query $Q_{text}$, we first pool the text embedding vectors of all the words to get the sentence-level feature, which accurately refers to the target instance in the video frame. Then, the sentence-level feature from the linguistic encoder is passed through an embedding layer to obtain 1-dim language token sequence.
   \begin{equation}
     Q_{\rm{text}} = Embed(Pool(f_{l}), W_e)
     \label{eq:q_text}
   \end{equation}
   where $W_e$ denotes learnable parameters of embedding layer.
   
   For the vision query $Q_{vis}$, 
   we know that two corner points $(x_1, y_1, x_2, y_2)$, or the center point plus the height and width $(cx, cy, w, h)$, define a bounding box.
   Given a sequence of bounding boxes in corner point format (similarly for another box format), we use \cref{eq:serialize} to quantify the coordinates of bounding box.
   \begin{equation}
     \widetilde{x}_{i} = round ( \frac{ x_{i} }{s} \times K )
     \quad
     \widetilde{y}_{i} = round ( \frac{ y_{i} }{s} \times K )
     \label{eq:serialize}
   \end{equation}
   where $x_{i}, y_{i}$ denote absolute coordinates at the search area level, $s$ is size of the search area, $K$ is the number of quantization bins, and $\widetilde{x}_{i}, \widetilde{y}_{i}$ represent quantized coordinates of bounding box. Following that, we serialize the quantized coordinates into multiple visual token sequences through an embedding layer.

   Once both $Q_{text}$ and $Q_{vis}$ are serialized, we concatenate them into a unified conditional queries.
   In this way, the built conditional queries can focus on the desired target object from different perspectives of appearance and high-level semantics.
   This query generation mechanism is more flexible to multi-modal scenarios and more robust to tracking the target instance in a video.
   Notably, unlike other queries with random initialization \cite{stark}, 
   the proposed conditional queries contain cues from multi-modal sources, which is crucial for learning VL tracking task.

\subsection{Task-agnostic Sequence Predictor}
   For the prediction head design of previous methods \cite{SiamRPN++,transt,stark}, most of them design task-specific prediction head for each task, i.e., classification head, regression head, and corner head.
   Each sub-task head requires its own optimization objectives that need to be learned independently.
   In the multi-tasks learning way, the weights of multiple loss functions also need to be tuned to fit sub-tasks.
   This increases the complexity of VL tracking modeling, making them highly customized and having limited generalization capabilities.

\begin{figure}[t]
      \centering
       \includegraphics[width=0.85\linewidth]{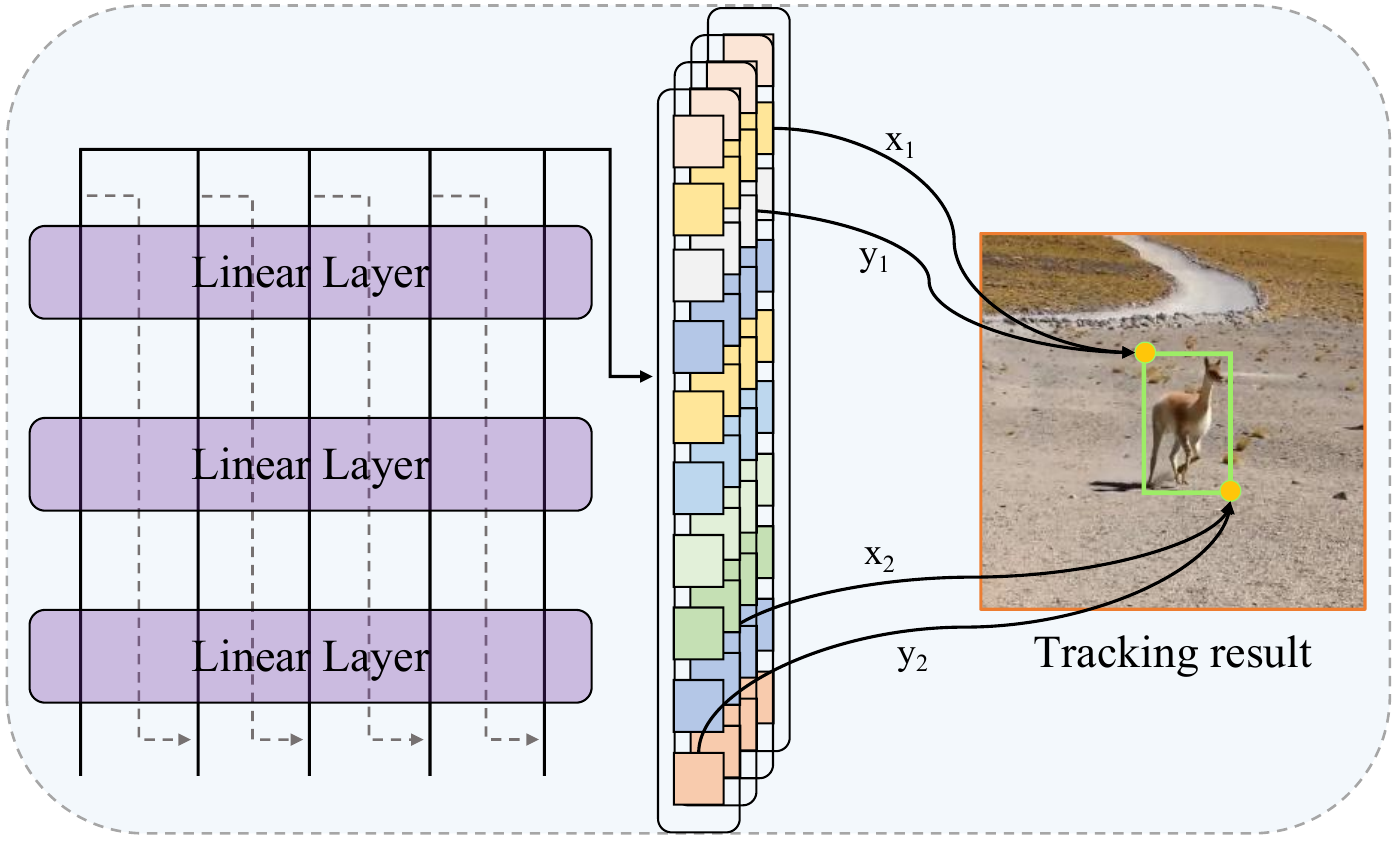}
       \caption{The architecture of task-agnostic sequence predictor. It iteratively generates the coordinate sequences describing the target instances in an auto-regressive manner.}
       \label{fig:head}
   \end{figure}

   Thus, we design a task-agnostic sequence predictor, that generates a sequence of attributes to describe the target instance, as shown in \cref{fig:head}.
   To be specific, three consecutive linear layers are added on top of the multi-modal decoder to further learn coordinate tokens.
   The prediction head outputs the probability of coordinate tokens sequences, where the indexes of the top four max scores represent the target localization in the current frame.
   By doing so, the model is able to escape from the guidance of traditional classifiers and reduce the design complexity of the prediction head.

   \textbf{Differences with the multi-task learning.}
   Most previous VL trackers combine multiple learning tasks to train a complex VL tracking model, i.e., jointly learning classification, regression and grounding tasks. However, instead of using the same multi-task learning strategy, we treat VL tracking as a token generation task. In this new paradigm, we reduce the learning effort of the model while avoiding the multi-task misalignment problem (i.e., the predicted bounding box with high classification score may not have high regression accuracy \cite{siamrcr}).
   Thus, our proposed method can greatly reduce the complexity of VL tracking modeling.

\subsection{Training and Inference}
   During training, since the training objective can be simplified, we use only cross-entropy loss to optimize MMTrack and realize token generation task.
   As a result, we no longer need to fine-tune the weights of specific loss functions to relieve training costs.
   
   During inference, we use two tokens to indicate the beginning and end of the target sequence prediction.
   In particular, the language query represents the start of VL tracking task execution, while the EOS token is regarded as the end of the token generation.
   After obtaining the coordinates of the target instance in an auto-regressive manner, the coordinates are transformed to obtain the final result.
   The inference pipeline only includes a forward network and no additional post-processing techniques.

   \begin{table*}[t]
    \centering
    \caption{Comparison with state-of-the-arts on four popular benchmarks: TNL2K \cite{tnl2k}, LaSOT \cite{lasot}, LaSOT$_{\rm{ext}}$ \cite{lasot-ext} and OTB99-Lang \cite{trackNL}.
    The vision-only type of methods are evaluated by bounding box initialization, while the vision-language (VL) type of methods are evaluated by joint bounding box and natural language initialization.
    It is worth noting that VLT$_{\rm{TT}}^{*}$ is evaluated on the local machine through its raw results.
    The best two results are highlighted in {\color{red}red} and {\color{blue}blue}, respectively.}
    \label{tab:results}
    \resizebox{\textwidth}{!}{
    \begin{tabular}{c|l|c|ccc|ccc|ccc|cc}
    \toprule
     \multicolumn{1}{c|}{\multirow{2}{*}{Type}} & \multicolumn{1}{c|}{\multirow{2}{*}{Method}} & \multicolumn{1}{c|}{\multirow{2}{*}{Source}}
      & \multicolumn{3}{c|}{TNL2K} &\multicolumn{3}{c|}{LaSOT} & \multicolumn{3}{c|}{LaSOT$_{\rm{ext}}$} & \multicolumn{2}{c}{OTB99-Lang} \\ \cline{4-14}
      & & & AUC & P${_{\rm{Norm}}}$ & P & AUC & P${_{\rm{Norm}}}$ & P & AUC & P${_{\rm{Norm}}}$ & P & AUC & P \\
      \midrule
      \multicolumn{1}{c|}{\multirow{15}{*}{\rotatebox{90}{Vision-only}}} & SiamFC \cite{SiamFC} & ECCV2016 & 29.5 & 45.0 & 28.6 & 33.6 & 42.0 & 33.9 & 23.0 & 31.1 & 26.9 & - & -\\
      & SiamRPN++ \cite{SiamRPN++} & CVPR2019 & 41.3 & 48.2 & 41.2 & 49.6 & 56.9 & 49.1 & 34.0 & 41.6 & 39.6 & - & - \\
      & DiMP \cite{DiMP50} & ICCV2019 & 44.7 & 51.3 & 43.4 & 56.9 & 65.0 & 56.7 & 39.2 & 47.6 & 45.1 & - & - \\
      & SiamBAN \cite{Siamban} & CVPR2020 & 41.0 & 48.5 & 41.7 & 51.4 & 59.8 & 52.1 & - & - & - & - & - \\
      & TransT \cite{transt} & CVPR2021 & 50.7 & 57.1 & 51.7 & 64.9 & 73.8 & 69.0 & - & - & - & - & - \\
      & Stark \cite{stark} & ICCV2021 & - & - & - & 67.1 & 77.0 & - & - & - & - & - & - \\
      & KeepTrack \cite{keeptrack} & ICCV2021 & - & - & - & 67.1 & 77.2 & 70.2 & 48.2 & - & - & - & - \\
      & BeamTracking \cite{xwang2} & TIP2022 & - & - & - & 63.9 & 67.1 & 61.5 & - & - & - & - & - \\
      & GTELT \cite{GTELT} & CVPR2022 & - & - & - & 67.7 & - & - & 45.0 & 54.2 & 52.2 & - & - \\
      & SBT-B \cite{SBT} & CVPR2022 & - & - & - & 65.9 & - & 70.0 & - & - & - & - & - \\
      & Mixformer \cite{mixformer} & CVPR2022 & - & - & - & 69.2 & 78.7 & 74.7 & - & - & - & - & - \\
      & TransInMo \cite{TransInMo} & IJCAI2022 & 52.0 & 58.5 & 52.7 & 65.7 & 76.0 & 70.7 & - & - & - & - & - \\
      & OSTrack-256 \cite{ostrack} & ECCV2022 & 54.3 & - & - & 69.1 & 78.7 & 75.2 & 47.4 & 57.3 & 53.3 & - & - \\
      & OSTrack-384 \cite{ostrack} & ECCV2022 & 55.9 & - & - & 71.1 & 81.1 & 77.6 & 50.5 & 61.3 & 57.6 & - & - \\
      & AiATrack \cite{aiatrack} & ECCV2022 & - & - & - & 69.0 & 79.4 & 73.8 & 47.7 & 55.6 & 55.4 & - & - \\
      & SimTrack \cite{simtrack} & ECCV2022 & - & - & - & 69.3 & 78.5 & - & - & - & - & - & - \\
      & SeqTrack \cite{seqtrack} & CVPR2023 & 56.4 & - & - & 71.5 & 81.1 & 77.8 & 50.5 & 61.6 & 57.5 & - & - \\
      \hline
      \multicolumn{1}{c|}{\multirow{12}{*}{\rotatebox{90}{Vision-Language}}} & Li-II \cite{trackNL} & CVPR2017 & - & - & - & - & - & - & - & - & - & 55.0 & 72.0 \\
      & Wang \cite{DATrack} & arXiv2018 & - & - & - & 27.7 & - & 30.4 & - & - & - & 65.8 & 89.1 \\ 
      & Feng \cite{NL-RPN} & arXiv2019 & 25.0 & 34.0 & 27.0 & 50.0 & - & 56.0 & - & - & - & 67.0 & 73.0 \\ 
      & Feng \cite{lstmtrack} & WACV2020 & 25.0 & 33.0 & 27.0 & 35.0 & - & 35.0 & - & - & - & 61.0 & 79.0 \\ 
      & SNLT \cite{SNLT} & CVPR2021 & - & - & - & 54.0 & 63.6 & 57.4 & - & - & - & 66.6 & 84.8 \\ 
      & GTI \cite{GTI} & TCSVT2021 & - & - & - & 47.8 & - & 47.6 & - & - & - & 58.1 & 73.2 \\ 
      & TNL2K-II \cite{tnl2k} & CVPR2021 & 42.0 & 50.0 & 42.0 & 51.3 & - & 55.4 & - & - & - & 68.0 & 88.0 \\ 
      & Li \cite{cmtrack} & CVPRW2022 & 44.0 & 52.0 & 45.0 & 53.0 & 56.0 & - & - & - & - & 69.0 & 91.0 \\ 
      & VLT$_{\rm{TT}}^{*}$ \cite{VLT} & NeurIPS2022 & 54.7 & 71.8  & 55.3 & {\color{blue}67.3} & {\color{blue}80.2} & {\color{blue}71.5} & {\color{blue}48.4} & {\color{red}59.9} & {\color{blue}54.3} & {\color{red}74.0} & 89.8 \\ 
      & TransVLT \cite{xwang1} & PRL2023 & 56.0 & 61.7 & - & 66.4 & - & 70.8 & - & - & - & 69.9 & {\color{blue}91.2} \\ 
      & JointNLT \cite{jointNLT} & CVPR2023 & {\color{blue}56.9} & {\color{blue}73.6} & {\color{blue}58.1} & 60.4 & 69.4 & 63.6 & - & - & - & 65.3 & 85.6 \\
      & \textbf{MMTrack} & Ours & {\color{red}58.6} & {\color{red}75.2} & {\color{red}59.4} & {\color{red}70.0} & {\color{red}82.3} &{\color{red}75.7} & {\color{red}49.4} & {\color{red}59.9} & {\color{red}55.3} & {\color{blue}70.5} & {\color{red}91.8} \\ 
    \bottomrule
    \end{tabular} }
\end{table*}

\section{Experiments}
\label{sec:experiments}

\subsection{Implementation Details}
   \textbf{Model.} 
   We use RoBERTa-Base \cite{RoBERTa} as our linguistic encoder, and the pre-trained OSTrack \cite{ostrack} to initialize our visual encoder.
   The feature channels $C$ and $d$ are $768$ and $256$, respectively.
   The quantization bins $K$ is $1000$.
   For each input image pair, including a template patch with $192 \times 192$ pixels and a search patch with $384 \times 384$ pixels.

   \textbf{Training.}
   We adopt the training splits of LaSOT \cite{lasot}, TNL2K \cite{tnl2k}, RefCOCOg-google \cite{refcocog-google}, and OTB99-Lang \cite{trackNL} as the training set. $60,000$ image pairs are randomly sampled in each epoch.
   We employ the AdamW \cite{adamw} to optimize the network parameters with initial learning rate of $4 \times 10^{-5}$ for the backbone, $4 \times 10^{-4}$ for the rest, and set the weight decay to $10^{-4}$.
   We set the training epochs to 150 epochs. The learning rate drops by a factor of 10 after 125 epochs.
   The model is conducted on a server with two RTX-2080Ti GPUs and set the batch size to be 32.
   In the inference phase, the tracker is tested on a RTX-2080Ti, and it runs at 36 FPS.
   The code is available at \href{https://github.com/Azong-HQU/MMTrack}{https://github.com/Azong-HQU/MMTrack}.

\subsection{Comparison with State-of-the-art}
   In this section, we evaluate the proposed model on four VL tracking benchmarks, including LaSOT \cite{lasot}, LaSOT$_{\rm{ext}}$ \cite{lasot-ext}, TNL2K \cite{tnl2k}, and OTB99-Lang \cite{trackNL}, and compare with existing state-of-the-art trackers.
   All experiments are preformed by using joint language and bounding box initialization.

   \textbf{TNL2K.} The TNL2K \cite{tnl2k} is a new large-scale benchmark for tracking by natural language task. Specifically, this dataset includes 2k video sequences and split 1300/700 for the training and testing, respectively. 
   Each video provides a sentence and corresponding bounding box to indicate the target object and its location.
   As shown in \cref{tab:results}, compared with two types of SOTA tracking methods (including vision-only tracking methods and VL tracking methods), the proposed framework achieves the best results and greatly outperforms the other methods.
   Compared with the vision-only tracking algorithms OSTrack-256 and OSTrack-384 \cite{ostrack}, our approach achieves 4.3\% and 2.7\% gains in terms of success score, respectively. This validates the importance of our multi-modal token unified learning strategy.
   Meanwhile, the MMTrack performs significantly better than the SOTA VL tracking algorithm VLT$_{\rm{TT}}$ \cite{VLT}, outperforming it by 3.9\%, 3.4\%, and 4.1\% in terms of success, normalized precision, and precision score, respectively.
   These results show that the use of linguistic information helps the model to learn semantic information about the target instance and greatly enhances target localization in complex scenarios.

\begin{figure*}
    \begin{center}
    \includegraphics[height=4.5cm]{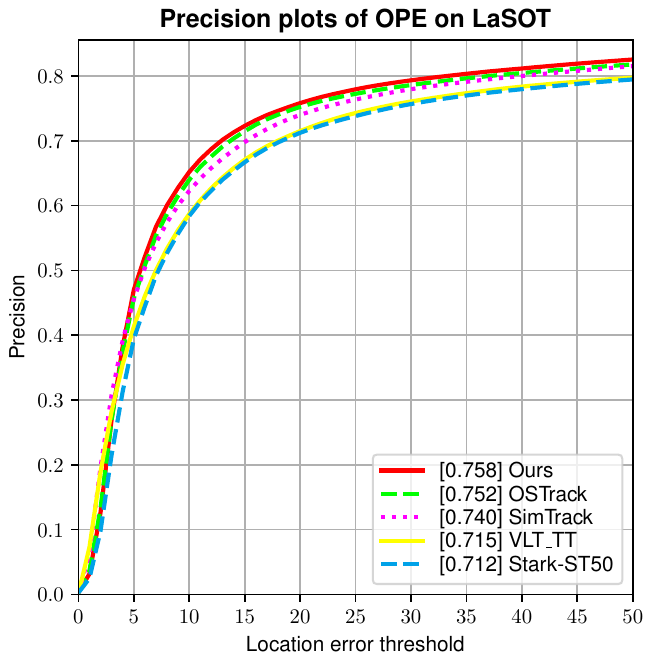}
    \includegraphics[height=4.5cm]{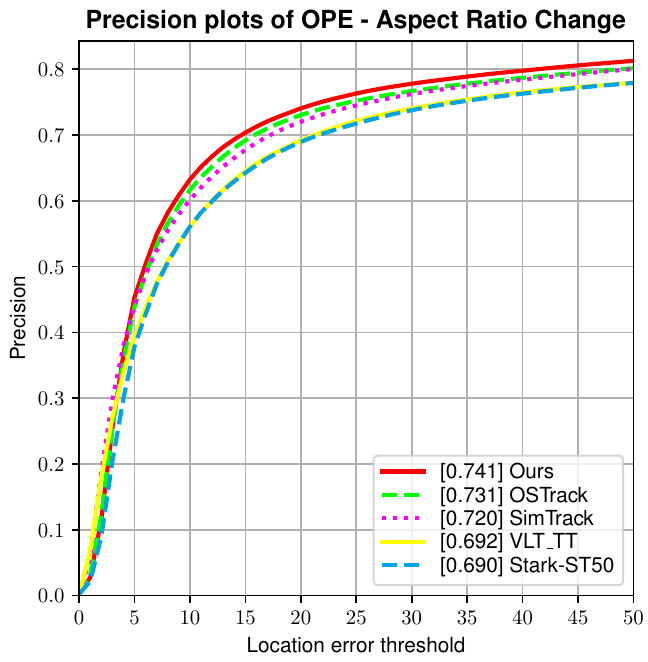}
    \includegraphics[height=4.5cm]{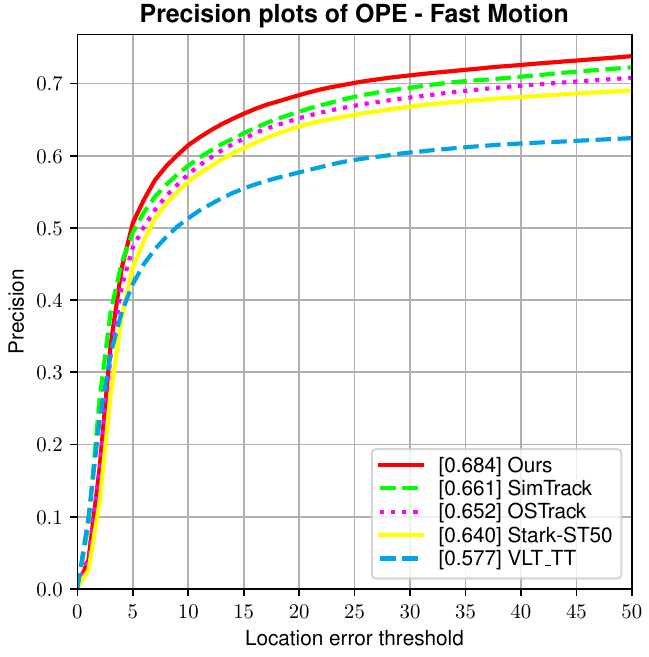}
    \includegraphics[height=4.5cm]{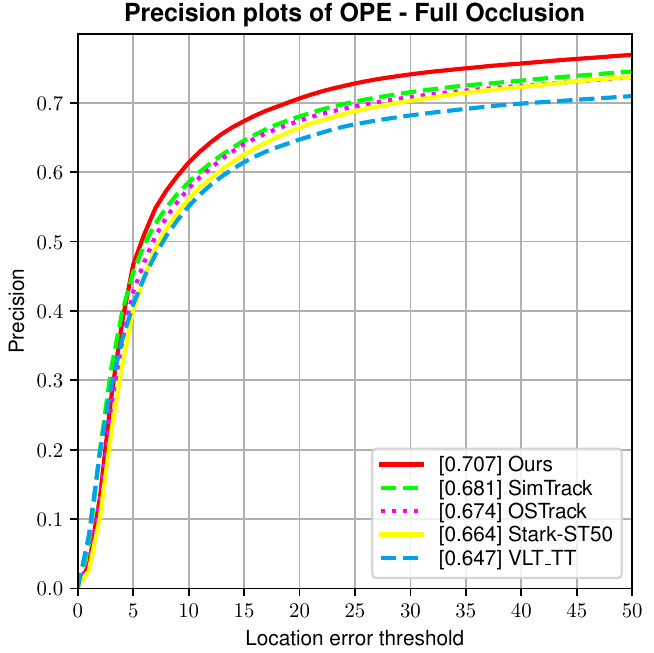}
    (a)\hspace{12em}(b)\hspace{12em}(c)\hspace{12em}(d)
    
    \includegraphics[height=4.5cm]{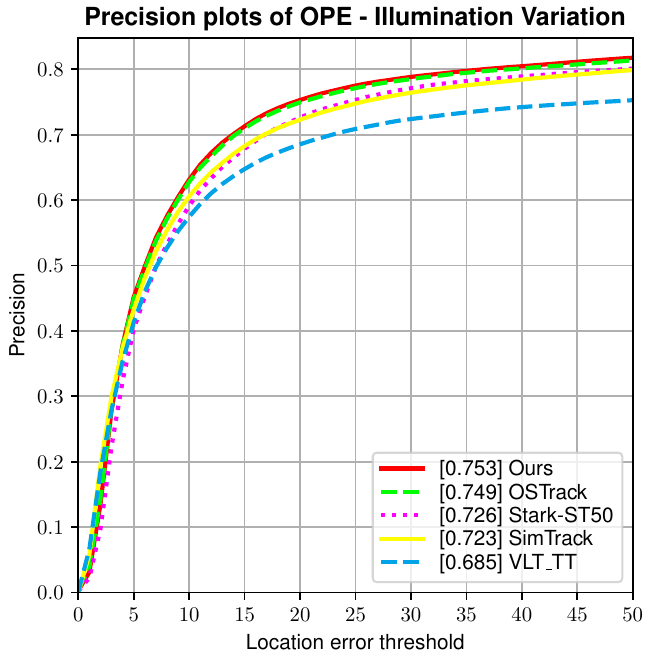}
    \includegraphics[height=4.5cm]{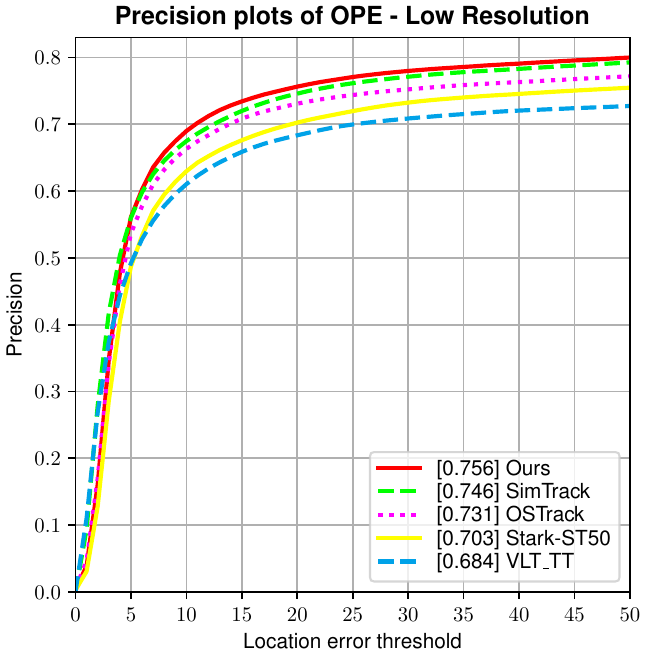}
    \includegraphics[height=4.5cm]{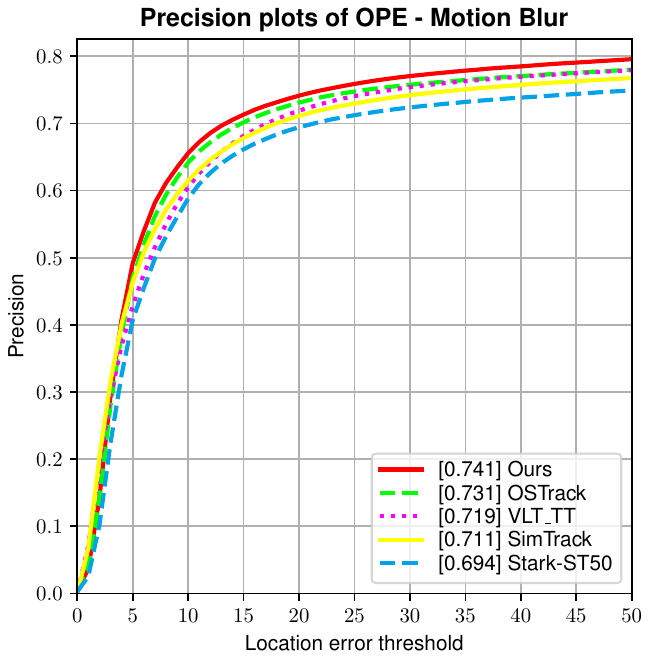}
    \includegraphics[height=4.5cm]{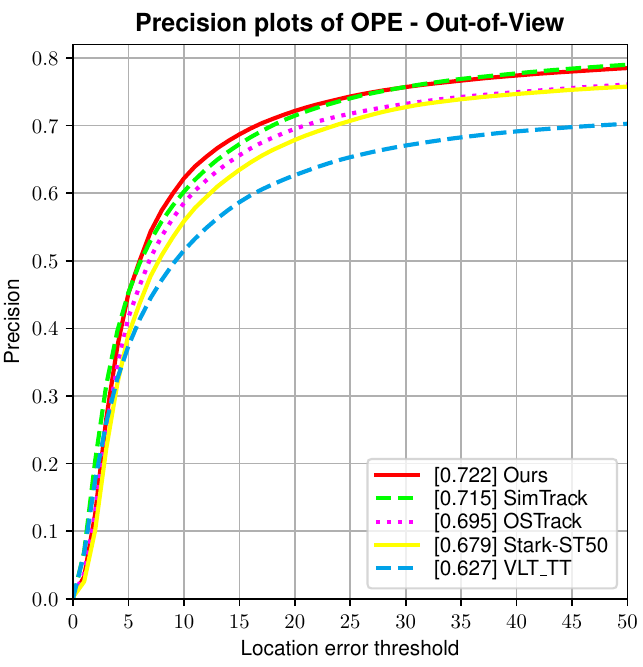}
    (e)\hspace{12em}(f)\hspace{12em}(g)\hspace{12em}(h)
    
    \includegraphics[height=4.5cm]{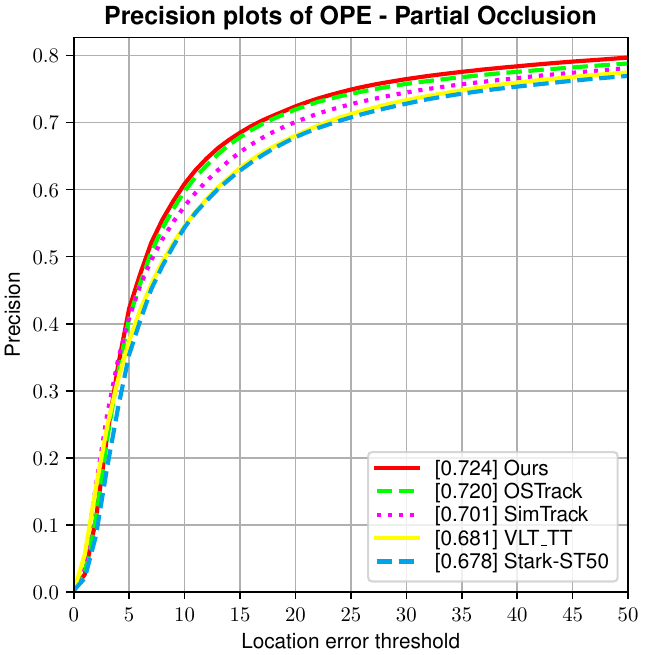}
    \includegraphics[height=4.5cm]{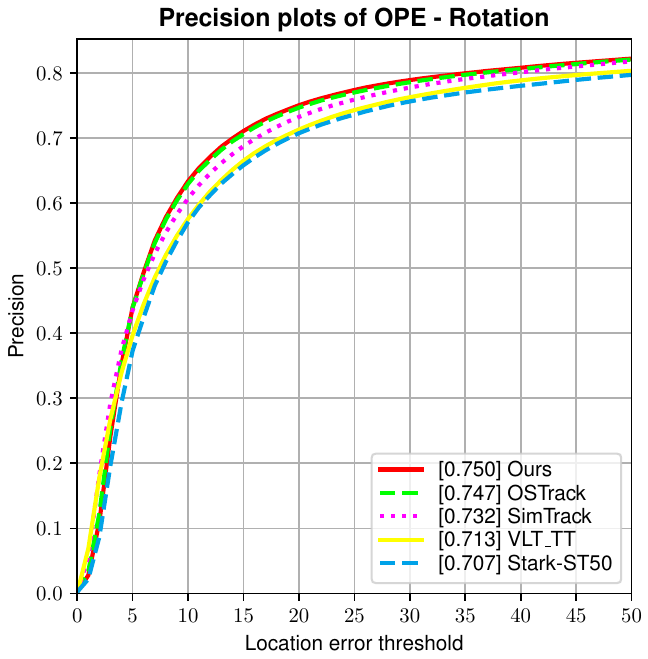}
    \includegraphics[height=4.5cm]{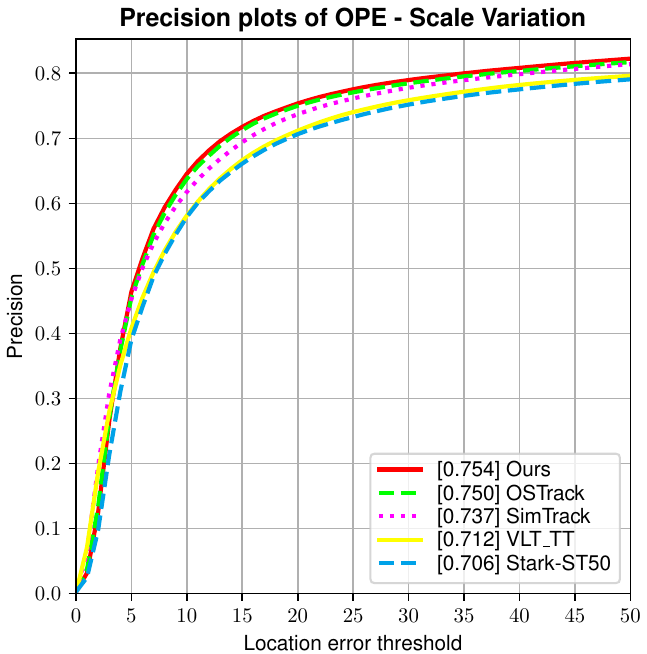}
     \includegraphics[height=4.5cm]{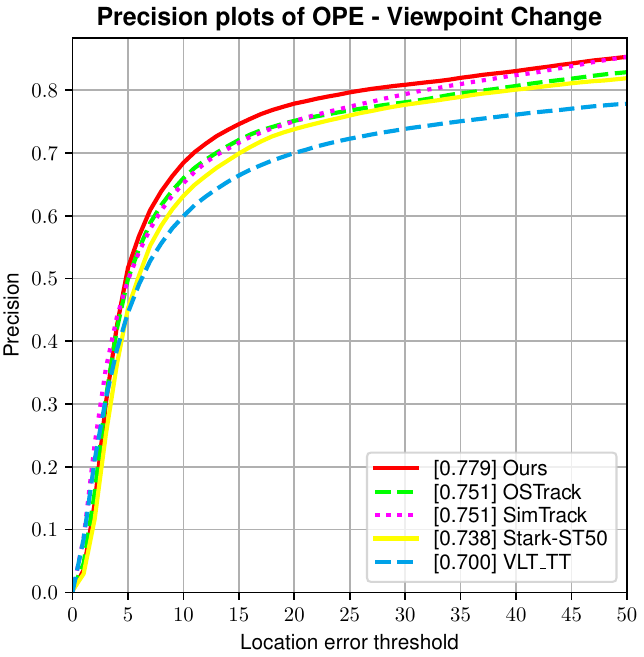}
    (i)\hspace{12em}(j)\hspace{12em}(k)\hspace{12em}(l)
\end{center}
\vspace{-3mm}
\caption{The precision plots with several attributes on LaSOT dataset \cite{lasot}. Such as aspect ratio change, fast motion, full/partial occlusion, illumination variation, low resolution, motion blur, out-of-view, rotation, scale variation, and viewpoint change.}
\label{fig:attrs}
\end{figure*}

\begin{figure*}
\begin{center}
\includegraphics[width=1\linewidth]{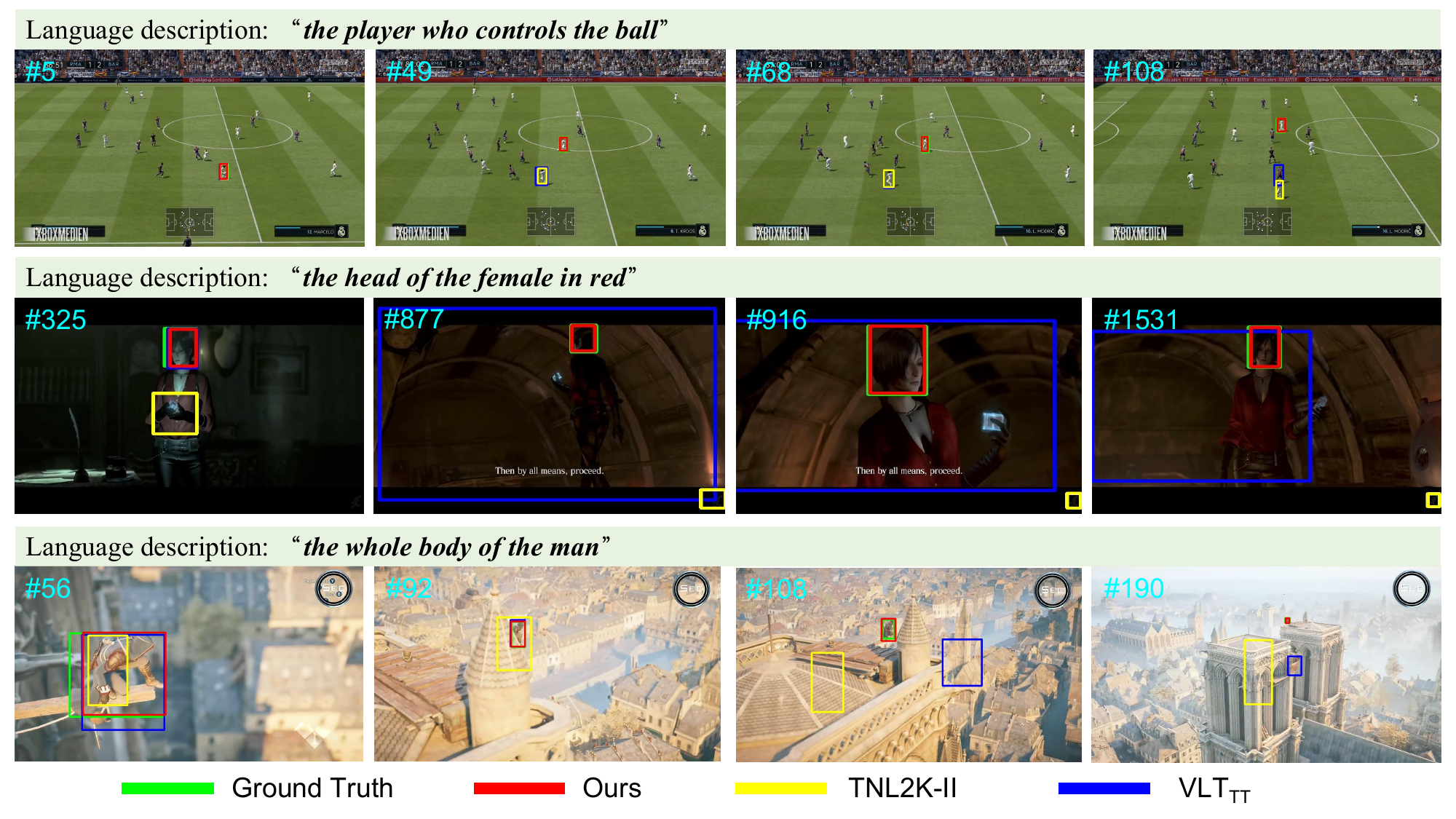}
\end{center}
\vspace{-5mm}
\caption{Qualitative comparison results of our tracker with other two VL trackers (i.e. $\rm{VLT}_{TT}$ \cite{VLT} and TNL2K-II \cite{tnl2k}) on three challenging sequences from the TNL2K benchmark. Better viewed in color with zoom-in.}
\label{fig:visual}
\end{figure*}

\begin{figure*}
\begin{center}
\includegraphics[width=1\linewidth]{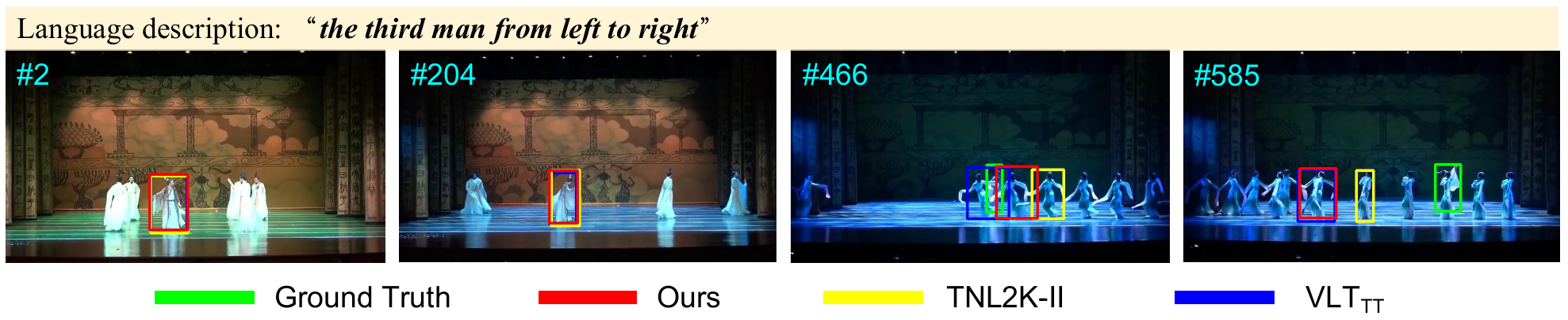}
\end{center}
\vspace{-5mm}
\caption{Failure case on a challenging sequence from TNL2K benchmark \cite{tnl2k}. Our tracker performs poorly with similar distractors and "long-term" occlusion (over 280 frames), which also pose great challenges to the compared trackers ($\rm{VLT}_{TT}$ \cite{VLT} and TNL2K-II \cite{tnl2k}). Better viewed in color with zoom-in.}
\label{fig:badcase}
\end{figure*}

   \textbf{LaSOT.} The LaSOT \cite{lasot} is a large-scale tracking benchmark that includes 1120 long-term videos for training and 280 for testing.   
   A natural language description with rich semantic information is provided for each sequence in this dataset.
   As shown in \cref{tab:results}, the proposed MMTrack achieves promising results among all vision-only and VL tracking algorithms.
   For example, compared with the vision-only tracking algorithm BeamTracking \cite{xwang2}, our approach achieves 6.1\%, 15.2\% and 14.2\% gains in terms of success, normalized precision and precision score, respectively.
   Compared with VL trackers TNL2K-II \cite{tnl2k} and Li \cite{cmtrack} with complex designs, our pipeline is simpler and surpasses them by a wide margin.
   Besides, the proposed method also improves 42.3\% and 3.6\% in terms of success score compared to Wang\cite{DATrack} and TransVLT\cite{xwang1}, respectively.

   Furthermore, we validate the effectiveness of MMTrack in the cases of aspect ratio change, fast motion, full/partial occlusion, illumination variation, low resolution, motion blur, out-of-view, rotation, scale variation, and viewpoint change. As shown in \cref{fig:attrs}, we visualize precision plots with multiple attributes on LaSOT dataset. Our MMTrack achieves good attribute evaluation results, compared to other vision-only (Stark \cite{stark}, OSTrack \cite{ostrack}, and SimTrack \cite{simtrack}) and vision-language (VLT$_{\rm{TT}}$ \cite{VLT}) trackers. These results demonstrate the ability of our tracker to effectively capture unified representational information about visual and linguistic modalities.

   \textbf{LaSOT$_{\rm{ext}}$.} A recent addition to LaSOT, the LaSOT$_{\rm{ext}}$ \cite{lasot-ext} contains 150 challenging long-term videos from 15 object classes.
   Like LaSOT, LaSOT$_{\rm{ext}}$ contains corresponding linguistic description for each sequence. Meanwhile, these new sequences are extremely challenging due to the presence of many similar distractors, making tracking difficult.
   As shown in \cref{tab:results}, we observe that our MMTrack achieves comparable result of 49.4\%, 59.9\% and 55.3\% in terms of success, normalized precision and precision score.
   For example, the MMTrack performs slightly better compared to the VLT$_{\rm{TT}}$ algorithm, outperforming it in terms of success and precision score by 1\% and 1\%, respectively, while the normalized accuracy score is equal.
   These results are consistent with our expectation, that is, token learning mechanism based on visual and linguistic cues can effectively capture different modalities information, thereby improving tracking performance.

   On the other hand, the performance of our tracker on the LaSOT and LaSOT$_{\rm{ext}}$ datasets is yet slightly weaker than ostrack-384. After our careful analysis, the reason for this could potentially be attributed to the fact that both the LaSOT and LaSOText are long-term tracking datasets, where the target instance undergoes significant variations over time. However, the language annotations are solely based on the initial frames of the videos, rendering them unable to adapt to the prolonged dynamic variations in video content. Consequently, in long-term multi-modal tracking scenarios, similar to other VL tracking methods\cite{jointNLT,VLT}, our tracker is susceptible to the influence of the initial language description.

\begin{table*}[t]
    \centering
    \caption{Ablation studies of query construction, bounding box format and the number of quantization bins on the TNL2K \cite{tnl2k} benchmark. The best results are shown in the {\color{red}red} font.}
    \label{tab:study}
    \resizebox{\textwidth}{!}{
    \begin{tabular}{c|cc|cc|ccccc|ccc}
    \toprule
     \multicolumn{1}{c|}{\multirow{2}{*}{Num}} & 
     \multicolumn{2}{c|}{Query construction} & 
     \multicolumn{2}{c|}{Bounding box format} & 
     \multicolumn{5}{c|}{Quantization bins} & 
     \multicolumn{3}{c}{TNL2K} \\  \cline{2-13}
      & multi-cues & single-cue & corner & center & 50 & 100 & 500 & 1000 & 2000 & AUC & P${_{\rm{Norm}}}$ & P \\
      \midrule  
      1 & \checkmark & & \checkmark & & \checkmark & & & & & 51.7 & 71.3 & 43.2 \\
      2 & \checkmark & & \checkmark & & & \checkmark & & & & 55.2 & 73.4 & 54.5 \\
      3 & \checkmark & & \checkmark & & & & \checkmark & & & 58.5 & 74.8 & 59.1 \\ 
      4 & \checkmark &  & \checkmark &  &  &  &  & \checkmark & & {\color{red}58.6} & {\color{red}75.2} & {\color{red}59.4} \\
      5 & \checkmark & & \checkmark & & & & & & \checkmark & 58.3 & 74.6 & 59.2 \\
      6 & \checkmark & & & \checkmark & & & & \checkmark & & 56.1 & 74.5 & 58.0 \\
      7 &  & \checkmark & \checkmark &  &  &  &  & \checkmark & & 57.9 & 73.6 & 57.5 \\ 
    \bottomrule
    \end{tabular} }
\end{table*}

   \begin{table*}[t]
    \centering
    \caption{Comparison of different configurations of language-annotated data.
    The best two results are highlighted in {\color{red}red}.}
    \label{tab:dataset}
    \resizebox{\textwidth}{!}{
    \begin{tabular}{c|ccc|ccc|ccc|cc}
    \toprule
    \multicolumn{1}{c|}{\multirow{2}{*}{Exps}}
      & \multicolumn{3}{c|}{TNL2K} &\multicolumn{3}{c|}{LaSOT} & \multicolumn{3}{c|}{LaSOT$_{\rm{ext}}$} & \multicolumn{2}{c}{OTB99-Lang} \\ \cline{2-12}
      & AUC & P${_{\rm{Norm}}}$ & P & AUC & P${_{\rm{Norm}}}$ & P & AUC & P${_{\rm{Norm}}}$ & P & AUC & P \\
      \midrule
      \textcircled{1} & 57.6 & 74.2 & 58.7 & 62.7 & 73.0 & 66.5 & 40.4 & 49.4 & 43.4 & 67.7 & 88.7 \\
      \textcircled{2} & {\color{red}58.6} & {\color{red}75.2} & {\color{red}59.4} & {\color{red}70.0} & {\color{red}82.3} &{\color{red}75.7} & {\color{red}49.4} & {\color{red}59.9} & {\color{red}55.3} & {\color{red}70.5} & {\color{red}91.8} \\
    \bottomrule
    \end{tabular} }
\end{table*}

\begin{table}[t]
\centering
\caption{Comparison of performance, model parameters, and inference speed on TNL2K\cite{tnl2k}.}
\resizebox{\linewidth}{!}{
\begin{tabular}{l|cccc}
\toprule
Method & AUC & Parameters (M) & Speed (FPS) & Device \\
\midrule
VLT$_{\rm{TT}}$ \cite{VLT} & 54.7 & 100.9 & 35.48 & RTX-2080Ti \\
JointNLT\cite{jointNLT} & 56.9 & 153.0 & 25.59 & RTX-2080Ti \\
Ours & \textbf{58.6} & 176.9 & \textbf{36.18} & RTX-2080Ti \\
\bottomrule
\end{tabular} }
\label{tab:param}
\end{table}

   \textbf{OTB99-Lang.} The OTB99-Lang \cite{trackNL} is an extension of the OTB benchmark \cite{OTB2015} in terms of language description. 
   Specifically, the benchmark includes 99 challenging sequences and split 51/48 for the training and testing, respectively.
   The performance of trackers is reported using the area under the curve (AUC) and precision (P) metrics in accordance with the protocol established by OTB official evaluation.
   As shown in \cref{tab:results}, our approach produced results that were competitive, with success and precision scores of 70.5\% and 91.8\%, respectively.
   But compared with VLT$_{\rm{TT}}$ \cite{VLT}, the score of our method ranks second. This could be since the annotations in OTB benchmark are relatively rough, which causes our tracker to perform unevenly.

\subsection{Ablation Study}
   To explore the role of the various components in the proposed framework, we conduct comprehensive ablation studies on the TNL2K \cite{tnl2k} benchmark.

   \textbf{Study on the number of quantization bins.} As described in \cref{sec:query}, the proposed conditional query is a crucial component in our MMTrack framework.
   The number of quantization bins determines the degree of refinement of the conditional query.
   As shown in \cref{tab:study}, we investigate the impact of the quantization bins on the tracking performance.
   It shows that increasing the number of quantization bins can significantly improve the tracking performance of our MMTrack.
   When the quantization bins gradually increases from 50 to 1000, the success score (AUC), normalized precision score (P$_{\rm{Norm}}$) and precision score (P) are improved by 6.9\%, 3.9\% and 16.2\%, respectively.
   It is worth noting that too many quantization bins cannot improve tracking performance, but rather converge to a stable performance.
   These results are consistent with our expectation that the conditional queries have been quantified to a fine enough level.

   \textbf{Study on the format of bounding box.} Since the representation of bounding box is available in various formats, such as corner points format $(x_1, y_1, x_2, y_2)$ and the center point plus the height and width $(cx, cy, w, h)$, we conduct ablation experiments to compare the effects of different formats.
   As shown in \cref{tab:study}, compared to \#6, the setting of corner point format (\#4) obtains a more competitive tracking performance, outperforming by 2.5\%, 0.7\% and 1.4\% in terms of AUC, P$_{\rm{Norm}}$ and P, respectively.
   We believe that the prediction of center point plus width and height may be more difficult since the model has to learn both the representation of center point and the scale of target object.

   \textbf{Study on query construction based multi-cues or single-cue.} 
   In our proposed framework, query construction plays an outstanding role. 
   We perform comparative experiments to verify the effect of queries constructed with different cues on tracking performance.
   For the settings of query sequence, \emph{multi-cues} include language and bounding box information, while \emph{single-cue} includes only the bounding box information of the target.
   As shown in \cref{tab:study}, compared to \#7, the tracking performance of \#4 with multi-cues outperforms 0.7\%, 1.6\% and 1.9\% in AUC, P$_{\rm{Norm}}$ and P, respectively.
   These results demonstrate that one benefit of our tracker comes from the query construction based multi-cues, which enhances the model's target semantic perception.

   \textbf{Study on the importance of language-annotated data.} 
   Furthermore, we explore the impact of the different settings of language-annotated data, i.e., \textcircled{1} Using only the TNL2K dataset to train our model, \textcircled{2} Using the full dataset containing LaSOT, RefCOCOg-google, TNL2K, and OTB99-Lang to train our model.
   As shown in \cref{tab:dataset}, compared to the tracking results trained with only TNL2K dataset (\textcircled{1}), we can find that the experimental results of \textcircled{2} achieve a better performance. For example, \textcircled{1} achieves good scores on the TNL2K dataset, but its performance severely degraded when evaluated on other datasets. These results demonstrate that language-annotated data is crucial for vision-language tracking task. Insufficient language-annotated data can lead to a degradation in performance. Therefore, our work aims to provide a multi-modal tracking solution to the vision-language tracking community, advancing research on language-driven multi-modal tracking algorithms.

\subsection{Analysis and Limitations}

\textbf{Speed Analysis.} As shown in \cref{tab:param}, we conduct a complete evaluation of execution efficiency, including model performance, inference speed and model parameters on TNL2K benchmark.
Our MMTrack tracker has approximately 176.9 million (M) parameters. Although our model has more parameters than VLT$_{\rm{TT}}$ \cite{VLT} and JointNLT \cite{jointNLT}, it does not affect our inference speed and model performance. This is mainly attributed to two factors. Firstly, inference speed and model parameters are not directly related, and may also depend on network design and execution process. For example, our linguistic encoder has a large number of parameters (85.805 M), but it only needs to be executed once in each video inference, so it has very little effect on the inference speed. Secondly, from an algorithmic perspective, since the VL tracking algorithm only needs to track a single target instance, our framework adopts a short token sequence for modeling, which does not significantly affect the model's inference speed. On the other hand, the proposed MMTrack outperforms VLT$_{\rm{TT}}$ and JointNLT in terms of both inference speed and model performance, yielding impressive results.

\textbf{Visualization.} To intuitively demonstrate the effectiveness of the proposed method, especially in complex scenarios such as target instance changes, background clutter, and scale variation, we visualize the tracking results of our MMTrack and two advanced VL trackers (such as VLT \cite{VLT} and TNL2K-II \cite{tnl2k}) on TNL2K benchmark. As shown in \cref{fig:visual}, due to the high-level semantics provided to enhance the unified VL representation, our tracker far outperforms the latest VL trackers VLT and TNL2K-II on these video sequences.

\textbf{Limitations.} Furthermore, we discuss the limitations of our algorithm. In vision-language tracking scenarios, multi-modal trackers still suffer from the challenges of similar distractors and "long-term" occlusion. As shown in \cref{fig:badcase}, the third actor (target object) from left to right is occluded for more than 280 frames and interfered by other similar actors, which prevents our tracker from accurately learning the target features. One potential solution for these complex challenges is to design a target association strategy with multi-frames.

\section{Conclusion}
\label{sec:conlusion}
   Vision-Language Tracking is a crucial area in computer vision. In this work, we reformulate VL tracking as a token generation task and propose a new VL tracking pipeline from a unified modeling perspective. Furthermore, the proposed MMTrack avoids sensitive loss functions and sub-tasks learning, simplifying multi-modal tracking modeling. Extensive experiments show that the proposed MMTrack outperforms previous methods on four VL tracking benchmarks. We hope that this work will encourage more researchers to participate in the VL tracking task.

\section*{Acknowledgments}
This work was supported by the Project of Guangxi Science and Technology (No.2022GXNSFDA035079), the National Natural Science Foundation of China (No.61972167 and U21A20474), the Guangxi ”Bagui Scholar” Teams for Innovation and Research Project, the Guangxi Collaborative Innovation Center of Multi-source Information Integration and Intelligent Processing, the Guangxi Talent Highland Project of Big Data Intelligence and Application, and the Research Project of Guangxi Normal University (No.2022TD002).

\bibliographystyle{ieeetr}
\bibliography{ref}

\end{document}